\documentclass[conference]{IEEEtran}
\IEEEoverridecommandlockouts

\usepackage{cite}
\usepackage{amsmath,amssymb,amsfonts}

\usepackage{graphicx}
\usepackage{textcomp}
\usepackage{xcolor}
\usepackage{algorithm}
\usepackage[noend]{algpseudocode}
\usepackage[letterpaper, left=0.75in, right=0.75in, top=1in, bottom=1in]{geometry}

\def\BibTeX{{\rm B\kern-.05em{\sc i\kern-.025em b}\kern-.08em
    T\kern-.1667em\lower.7ex\hbox{E}\kern-.125emX}}

\begin{document}

\title{Multi-Objective Reinforcement Learning for Cognitive Radar Resource Management}

\author{\IEEEauthorblockN{Ziyang Lu,  Subodh Kalia, M. Cenk Gursoy,  Chilukuri K. Mohan,  Pramod K. Varshney} 
\IEEEauthorblockA{Department of Electrical Engineering and Computer Science, Syracuse University, Syracuse NY, 13066 
\\
\{zlu112, skalia, mcgursoy, ckmohan, varshney\}@syr.edu
}}



\maketitle

\begin{abstract}
The time allocation problem in multi-function cognitive radar systems focuses on the trade-off between scanning for newly emerging targets and tracking the previously detected targets. We formulate this as a multi-objective optimization problem and employ deep reinforcement learning to find Pareto-optimal solutions and compare deep deterministic policy gradient (DDPG) and soft actor-critic (SAC) algorithms. Our results demonstrate the effectiveness of both algorithms in adapting to various scenarios, with SAC showing improved stability and sample efficiency compared to DDPG. We further employ the NSGA-II algorithm to estimate an upper bound on the Pareto front of the considered problem. This work contributes to the development of more efficient and adaptive cognitive radar systems capable of balancing multiple competing objectives in dynamic environments.
\end{abstract}


\section{Introduction}

Cognitive radar systems \cite{haykin2006cognitive} have emerged as a promising technology for enhancing radar performance in complex and dynamic environments, and recent advances are summarized
in \cite{charlish2020development}, \cite{greco2018cognitive} and \cite{gurbuz2019overview}.
These  systems can adapt their operational parameters in real-time based on the current conditions, leading to improved detection and tracking performance. 
However, determining the optimal time allocation strategy
among various functions remains a key challenge in cognitive radar systems, due to 
conflicting objectives
\cite{charlish2020development}, 
especially 
when 
both wide-area surveillance 
and accurate tracking of multiple targets are important.
The authors in \cite{orman1996scheduling} proposed scheduling approaches for multifunction phased array radar systems, and 
the authors in \cite{deligiannis2017game} formulated the power allocation problem in a multi-radar system as a non-cooperative game and performed an analysis of the Nash equilibrium and its convergence. 

Recent works demonstrate the applicability of deep reinforcement learning (DRL) 
for adapting radar operational parameters to dynamic environments,
e.g., 
for radar detection and tracking in congested spectral environments 
\cite{thornton2020deep}. In \cite{stephan2022scene}, the authors employed DRL for scene-adaptive radar tracking. The work in \cite{durst2021quality} proposed a DRL approach for quality of service-based radar resource management, effectively balancing multiple performance metrics. The work in \cite{9455344} applied reinforcement learning to the revisit interval selection problem in multifunction radars, demonstrating that a Q-learning approach could achieve lower tracking load compared to conventional methods while maintaining comparable track loss probability. 


We formulate the radar resource allocation problem as a multi-objective optimization problem, addressing scanning vs. tracking.
Prior works attempt to
optimize a weighted sum of multiple objective functions with a fixed set of weights,
but this approach is inadequate for  real-world applications, where the importance of different objectives can vary depending on 
task requirements and operational environments. 
Instead,
we generate and analyze the Pareto front by adjusting the tradeoff coefficient between scanning and tracking objectives.
We evaluate two state-of-the-art deep RL algorithms, namely deep deterministic policy gradient (DDPG) \cite{lillicrap2015continuous} and soft actor-critic (SAC) \cite{haarnoja2018soft} to determine Pareto-optimal solutions.
We also use a well-known multi-objective evolutionary algorithm (NSGA-II) \cite{deb2002fast} to establish an upper bound on the Pareto front of the considered problem. We conduct a detailed analysis of the learned policies, demonstrating how resource allocation strategies adapt to different prioritizations of scanning and tracking objectives.

Section II details the system model of the considered radar time allocation problem. Section III describes the
considered methodologies, including the DDPG and SAC algorithms and
the process for generating the Pareto front. In Section V, we present and analyze the numerical results, and conclude in Section VI.

\section{System Model}

We consider a cognitive radar system that allocates its time between scanning for potential targets and tracking detected ones. The system model is based on our previous work \cite{10215369} and we summarize it below.

\subsection{Target Motion Model}
The state of a target at time $t$ is defined as $\mathbf{x}_t = [x_t, y_t, \dot{x}_t, \dot{y}_t]^T$, where $(x_t, y_t)$ is the target's location and $(\dot{x}_t, \dot{y}_t)$ are its horizontal and vertical velocities. The target motion is modeled as
\begin{equation}
    \mathbf{x}_{t+1} = \mathbf{F}_t \mathbf{x}_t + \mathbf{w}_t
\end{equation}
where $\mathbf{F}_t$ is the state transition matrix and $\mathbf{w}_t$ is Gaussian  maneuverability noise with covariance $\mathbf{Q}_t$.

\subsection{Measurement Model}
The radar obtains noisy measurements of the range $r$ and azimuth angle $\theta$ of a target:
\begin{equation}
	\mathbf{z_t} = h(\mathbf{x_t}) + \mathbf{v_t} = \left[\sqrt{x_t^2+y_t^2}, \quad \text{tan}^{-1}\left(\frac{y_t}{x_t}\right)\right]^T + \mathbf{v_t}
\end{equation}
where $\mathbf{v}_t$ is Gaussian measurement noise with covariance $\mathbf{R}_t$.

\subsection{Tracking Model}
We employ the extended Kalman filter (EKF), a nonlinear estimation technique for radar tracking applications \cite{bar2011tracking}. 
The tracking performance for each target at time $t$ is quantified by a cost function $c_t$, defined as\begin{equation}
    c_t(\tau_t) = \text{trace}(\mathbf{E}\mathbf{P}_{t|t}\mathbf{E}^T)
\end{equation}
where $\mathbf{P}_{t|t} \in \mathbb{R}^{4\times4}$ is the covariance matrix of the posterior state estimate, obtained by EKF. $\mathbf{E}$ is a projection matrix that extracts the position components from the state vector. $\tau_t$ is the dwell time, i.e., the duration in which the radar focuses on an individual target during a measurement cycle.
Increasing the dwell time 
improves the accuracy of state estimates and reduces uncertainty.
%
We fix the duration of the measurement cycle $T_t$ to be 
$T_0$. 
Increasing the dwell time $\tau_t$ for one target 
improves its tracking accuracy
at the expense of other targets and the scanning performance.

\subsection{Scanning Model}
A uniform circular array (UCA) radar is used for scanning. The signal-to-noise ratio (SNR) during scanning is given by
\begin{equation}
    \text{SNR}_\text{scan} = \frac{P_t \tau_\text{beam} G_t G_r \lambda_r^2 \sigma}{(4\pi)^3 r^4 L k T_s}
\end{equation}
where $\tau_\text{beam}$ is the beam duration, $P_t$ is the transmit power, $G_t$ and $G_r$ are the transmit and receive antenna gains, $\lambda_r$ is the radar signal wavelength, $\sigma$ is the radar cross section of the target, $r$ is the radar-target distance, $L$ is a loss factor, $k$ is Boltzmann's constant, and $T_s$ is the system temperature. $\tau_\text{beam}$ is formulated as
\begin{equation}
    \tau_{s} = \frac{360^\circ}{\phi}\tau_{beam}
\end{equation}
where $\tau_s$ denotes the total time allocated to scanning, 
$\phi$ is the phase delay between adjacent radar beams, and $\tau_{beam}$ is the time duration of each beam.

Given specified probabilities of detection ($P_d$) and false alarm ($P_f$), we can determine the minimum required SNR. Using this $\text{SNR}_\text{min}$, we can derive the maximum detectable range $r_\text{max}$ by setting $\text{SNR}_\text{scan} = \text{SNR}_\text{min}$ and solving for $r$:
\begin{equation}
    r_\text{max} = \left(\frac{P_t \tau_\text{beam} G_t G_r \lambda_r^2 \sigma}{(4\pi)^3 L k T_s \cdot \text{SNR}_\text{min}}\right)^{1/4}.
    \label{rmax}
\end{equation}

This $r_\text{max}$ represents the maximum range at which the radar can detect a target with the required $P_d$ and $P_f$. For distances $r \leq r_\text{max}$, the actual probability of detection will meet or exceed the required $P_d$ while maintaining the desired $P_f$.

To quantify the scanning performance of our cognitive radar system, we introduce the metric $\Gamma$, defined as the ratio of the maximum detectable area to a reference area:
\begin{equation}
    \Gamma = \frac{A_\text{max}}{A_\text{ref}} = \frac{\pi r_\text{max}^2}{\pi r_0^2} = \left(\frac{r_\text{max}}{r_0}\right)^2
\end{equation}
where $r_\text{max}$ is the maximum detectable range given in (\ref{rmax}). $r_0$ is a reference range, typically set to the default operating range of the radar. $A_\text{max} = \pi r_\text{max}^2$ is the maximum detectable area and $A_\text{ref} = \pi r_0^2$ is the reference area.

Increasing the time allocated for scanning results in an increase in
 $\Gamma$,  the capability of detecting a new target, its coverage region, and also detection probability at a given distance.

\subsection{Track Initialization}
Track initialization is the process in which a new target is captured by the radar system. The most common strategy for track initialization is the $M$-of-$N$ model. We use the $3$-of-$4$ track initialization model where a track is initialized if $3$ associated measurements/detections are obtained within $4$ successive scans. We define associated measurements using a global nearest neighbor (GNN) approach \cite{bar2011tracking}. Specifically, a measurement is considered associated with a previous one if the Euclidean distance between them falls below a predefined threshold.

\section{Problem Formulation}

In this section, we formulate the cognitive radar resource allocation problem as a constrained optimization task, balancing the performance of tracking of existing targets and scanning for new ones.

\subsection{Utility Function}

Using the linear scalarization method in multi-objective optimization, we define a utility function that combines tracking and scanning performances as follows:
\begin{equation}
    U_t(\{\tau_t^n\}_{n=1}^N) = -\sum_{n=1}^N c_t^n(\tau_t^n) + \beta \Gamma
    \label{utility}
\end{equation}
where
\begin{itemize}
    \item $c_t^n(\tau_t^n)$ is the tracking cost for target $n \in \{1,\ldots,N\}$ at time $t$, as defined in Section II.C;
    \item $\tau_t^n$ is the dwell time allocated to tracking target $n$;
    \item $\Gamma$ is the scanning metric as defined in Section II.D;
    \item $\beta$ is the tradeoff coefficient balancing the importance of scanning versus tracking.
\end{itemize}

\subsection{Optimization Problem}

Our goal is to find an optimal time allocation policy $\pi$ that maximizes the expected discounted sum of utilities over time, subject to a time budget constraint:
\begin{equation}
	\begin{aligned}
		\max_{\pi}  \quad & \sum_{m=0}^\infty \left[\gamma^m U_{t+m}^n(\tau_{t+m}^n)\right]\\	  
  \vspace{-.5cm}\textrm{s.t.} \quad & \sum_{m=0}^\infty\gamma^m\left(\sum_{n=1}^N \frac{\tau_{t+m}^n}{T_0} - \Theta_{max}\right) \leq 0.		\end{aligned}
	\label{formulated_constrained_opt}
\end{equation}
where $\gamma \in (0, 1]$ is the discount factor, $T_0$ is duration of the radar measurement cycle and $\Theta_{max}$ is the total time budget for tracking.

The constraint ensures that the total time allocated to tracking does not exceed a predefined budget, and the remaining time $\tau_s = T_0 - \sum_{n=1}^N \tau_t^n$ is allocated for the scanning task.

By introducing a dual variable $\lambda$, the problem can be relaxed to an unconstrained optimization problem:
\begin{align}
	\min_{\lambda_t \geq 0} \max_{\pi} \sum_{m=0}^\infty \gamma^m \Bigg[U_{t+m}^n(\tau_{t+m}^n)-\lambda_t\left(\sum_{n=1}^N \frac{\tau_{t+m}^n}{T_0} - \Theta_{max}\right)\Bigg].
	\label{dual}
\end{align}

\subsection{Pareto Front Generation}

To explore the trade-off between tracking and scanning performance, we vary the tradeoff coefficient $\beta$ and solve the optimization problem for each value. This approach allows us to generate the Pareto front, providing insights into the achievable combinations of tracking accuracy and scanning effectiveness.

For each $\beta$ value, we determine the optimal policy $\pi^*$ that maximizes the combined objective function. By varying $\beta$ and solving the optimization problem, we generate a set of Pareto-optimal solutions. This approach characterizes the trade-off frontier between tracking accuracy and scanning performance. The resulting Pareto front provides radar operators with a comprehensive view of achievable performance combinations, assisting them in selecting the most suitable operating points based on specific mission objectives and constraints.

\section{Constrained Deep Reinforcement Learning Framework}
A constrained deep reinforcement learning (CDRL) framework is proposed for solving the constrained optimization problem (\ref{formulated_constrained_opt}).

\subsection{State}
The state $s_t$ at time $t$ is defined as
\begin{equation}
    s_t = [\{c_{t-1}^n(\tau_{t-1}^n)\}_{n=1}^N, \{\tau_{t-1}^n\}_{n=1}^N, \lambda_{t-1}, \beta]
\end{equation}
where $\{c_{t-1}^n(\tau_{t-1}^n)\}_{n=1}^N$ are the tracking costs for all $N$ targets, $\{\tau_{t-1}^n\}_{n=1}^N$ are the dwell times allocated in the previous time slot, $\lambda_{t-1}$ is the dual variable, and $\beta$ is the tradeoff coefficient. The size of the state is $2N+2$.

\subsubsection{Action}
The action $a_t$ is the set of dwell times allocated to each target:
\begin{equation}
    a_t = \{\tau_t^n\}_{n=1}^N, \quad \text{where } \tau_t^n \in [0, T_0]
\end{equation}

\subsubsection{Reward}
To solve the unconstrained optimization problem (\ref{dual}), the reward function $r_t$ is defined as
\begin{equation}
    r_t = U_t(\{\tau_t^n\}_{n=1}^N) - \lambda_t\left(\sum_{n=1}^N \frac{\tau_t^n}{T_0} - \Theta_{max}\right)
\end{equation}
where $U_t(\{\tau_t^n\}_{n=1}^N)$ is the utility function defined in (\ref{utility}), and the second term is the penalty for violating the time budget constraint.

In the proposed CDRL framework, the dual variable $\lambda$ is updated simultaneously with the neural network parameters as
\begin{equation}
    \lambda_{t+1} = \max\left(0, \lambda_t + \alpha_\lambda \left(\sum_{n=1}^N \frac{\tau_t^n}{T_0} - \Theta_{max}\right)\right)
\end{equation}
where $\alpha_\lambda$ is the learning rate of the dual variable. The
update of $\lambda$ dynamically adjusts the penalty for constraint violations, guiding the learning process toward feasible solutions.

In this paper, we consider implementations with both DDPG and SAC algorithms for finding the Pareto front of our multi-objective radar resource allocation problem. While DDPG provides a deterministic policy that directly maps states to actions, SAC offers a stochastic policy that also considers maximizing an entropy term. This entropy maximization in SAC encourages exploration and improves robustness by learning a diverse set of effective policies. Specifically, SAC aims to maximize the expected return while also maximizing the entropy of the policy:
\begin{equation}
    J(\pi) = \mathbb{E}_{\tau \sim \pi} \left[\sum_{t=0}^{\infty} \gamma^t (r_t + \alpha H(\pi(\cdot|s_t)))\right]
\end{equation}
where $J(\pi)$ is the expected return of the policy $\pi$, $H(\pi(\cdot|s_t))$ is the entropy of the policy at state $s_t$, and $\alpha$ is a temperature parameter that balances reward maximization and entropy maximization. This formulation allows SAC to potentially find more diverse solutions along the Pareto front, especially in regions where multiple allocations might yield similar performance. Indeed, SAC is considered as an implementation of the \emph{concave-augmented Pareto Q learning algorithm} proposed in \cite{lu2023multiobjective}. By comparing DDPG and SAC, we aim to provide insights into the effectiveness of different CDRL implementations in solving complex, multi-objective radar resource allocation problems.

\section{Numerical Results and Analysis}

\subsection{Hyperparameters}
\vspace{-.2cm}
\begin{table}[!ht]
	\renewcommand{\arraystretch}{1}
	
	\caption{Simulation Parameters}
	\label{table1}
	\centering
	\small
	\begin{tabular}{|c||c|}
        \hline $N$ & 5\\
		\hline $\sigma_{r,0}^2$ ($m^2$)&16\\
		\hline $\sigma_{\theta,0}^2$ $(\text{rad}^2)$&1e-6\\
		\hline $\sigma_w$ ($(m/s^2)^2$) & 16\\
		\hline Measurement Cycle $T_0$ (s) & 2.5\\
        \hline Time Budget for Tracking ($\Theta_{max}$) & 0.9\\
        \hline Required Probability of Detection ($P_d$) & 0.9\\
        \hline Required Probability of False Alarm ($P_f$) & 1e-3\\
		\hline DRL Discount Factor $\gamma$ & 0.9\\
		\hline Initial Dual Variable ($\lambda_0$) & 5000\\
		\hline Step Size of Dual Variable ($\alpha_\lambda$) & 15000\\
        
		\hline
		
	\end{tabular}
	\label{table1}
\end{table}

The values of key hyperparameters are provided in Table \ref{table1}. In the DDPG implementation, critic networks have two feedforward layers, each consisting of 100 neurons, and the actor networks have two feedforward layers, each consisting of 256 and 128 neurons, respectively. The learning rates for all the networks are set to be 0.0001. In the SAC implementation, critic networks have two feedforward layers with 100 neurons each, while the actor network has two feedforward layers with 128 neurons each. In SAC, the learning rates for all networks are set to 0.0001 as well. Adam optimizer is employed during training. Simulations were conducted on a desktop computer equipped with an Intel Core i7-8700 CPU (3.20GHz, 6 cores, 12 logical processors). 

\subsection{Dynamic Radar Resource Management with CDRL }

\begin{figure}
    \centering
    \includegraphics[width=0.73\linewidth]{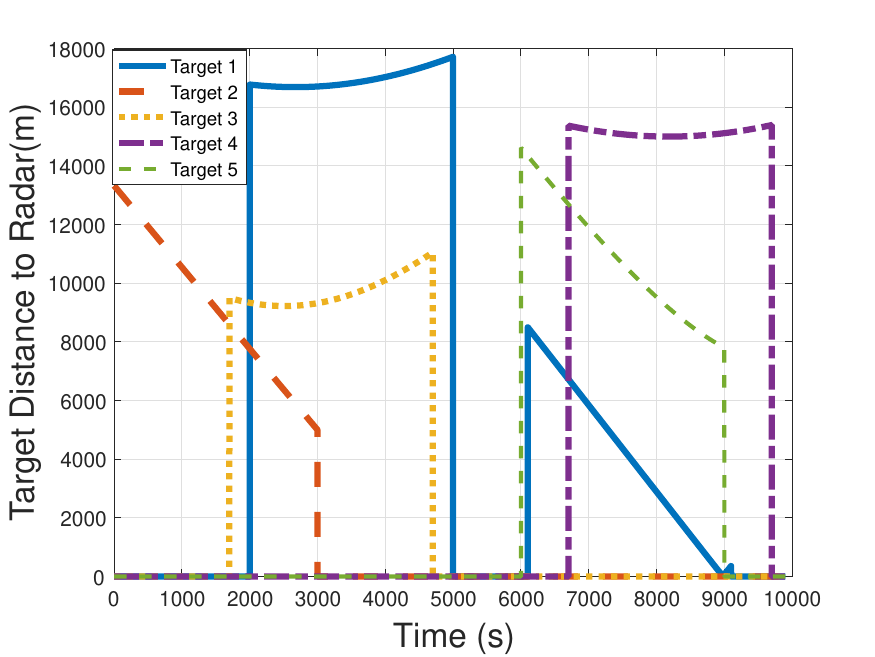}
    \caption{Distance of the Targets to the Radar}
    \label{dist}
\end{figure}

\begin{figure}
    \centering
    \includegraphics[width=0.75\linewidth]{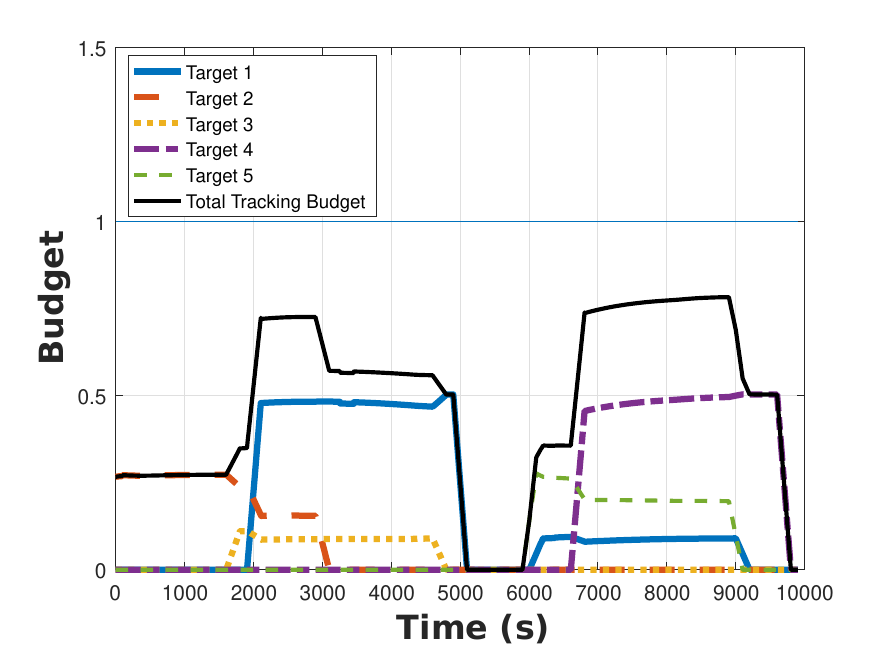}
    \caption{Time Allocation Strategy by CDRL-SAC}
    \label{budget}
\end{figure}

Fig. \ref{dist} shows the distance of the targets to the radar in an episode, where the targets are generated with random initial locations, velocities, and spawning time. The number of targets is also varied over time. 
For CDRL-SAC with $\beta=150000$, Fig. \ref{budget} illustrates that 
when more targets are tracked, the framework allocates more time to tracking tasks,
e.g., with three tracked targets when $t\in[7000, 9000]$.
CDRL-SAC also 
allocates more time to distant targets, 
which benefit from additional resources, 
e.g., during $t \in [7000, 9000]$, 
maximizing the overall reward function. 
This 
demonstrates the ability of the proposed CDRL-SAC framework to balance multiple objectives and adapt to dynamic radar environments.

\subsection{Comparison of Pareto Fronts}

We evaluate our approach using both DDPG and SAC algorithms,
comparing them with the equal allocation strategy (equally allocating the total tracking time to targets) as well as NSGA-II \cite{deb2002fast},
in terms of the aggregate tracking performance $obj_t = -\sum_{n=1}^N c_t^n(\tau_t^n)$ and scanning performance $obj_s = \Gamma$, while varying  the tradeoff coefficient $\beta \in [0, 300000]$.
Non-dominated  $(obj_t, obj_s)$ pairs constitute the Pareto front.
The NSGA-2 implementation uses 120 time allocation decision variables, (for scanning and for  tracking  5 potential targets  at 20 decision points) throughout the episode with 10000 time slots. 
The sum of time allocations at each decision point is constrained
to be $\leq T_0$. 
\begin{figure}[t]
    \centering
    \includegraphics[width=0.78\linewidth]{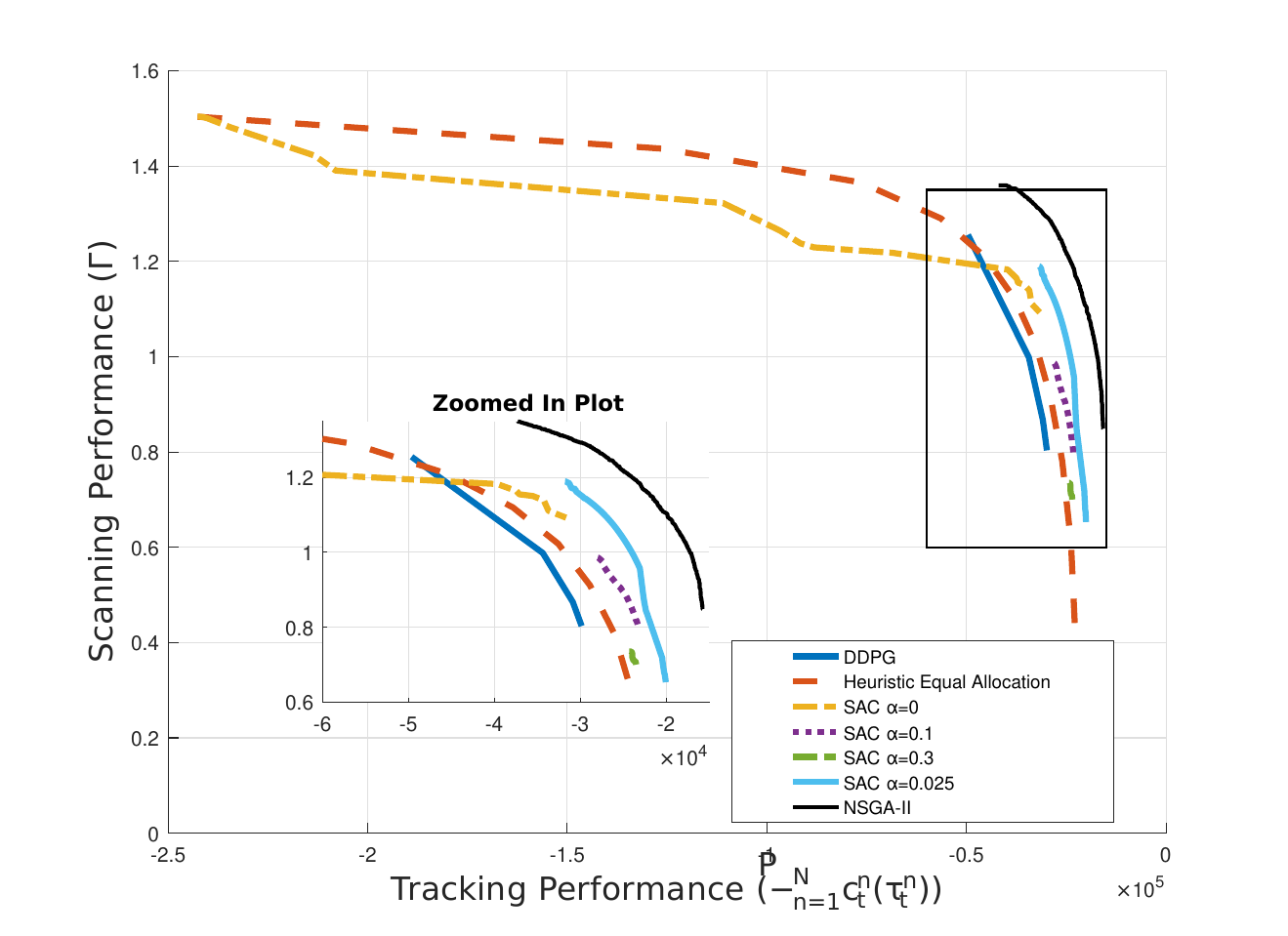}
    \caption{Comparison of Pareto Fronts}
    \label{pareto}
\end{figure}
Fig. \ref{pareto} compares the Pareto fronts achieved, 
resulting in the following observations:

\begin{itemize}
    \item 
    The increase in tracking performance generally leads to a decrease in scanning performance.

    \item CDRL-SAC with 
    $\alpha=0.025$ achieves the best Pareto front among all learning-based schemes. Its operation points 
    cover a wide range of the operational region and also consistently dominate 
    others. 

    \item 
    Excessive emphasis on entropy (i.e., higher $\alpha$) can lead to suboptimal performance 
    due to the over-prioritization of exploration over exploitation.

    \item Some exploration is necessary, as demonstrated by the inadequacy of CDRL-SAC with $\alpha=0$ and the DDPG algorithm, 
    which focus solely on reward maximization. 

    \item 
    With 60 different $\beta$ values, CDRL-SAC 
    required approximately 48 minutes to generate the Pareto front over the entire episode of $10000$ time steps. 
    
    \item 
     NSGA-II was implemented with tournament selection, simulated binary crossover (with $\eta = 15$ and probability 0.9), and polynomial mutation (with $\eta = 20$) 
     using a population size of 2400 over 1000 generations.
     Computations were distributed across ten nodes, each equipped with 128 cores. 
    NSGA-II provided useful upper bounds 
    on the achievable Pareto front, but 
had high computational demands (11.5 hours in our simulations).

\end{itemize}


\section{Conclusions}

In this paper, we proposed a constrained deep reinforcement learning (CDRL) framework for determining Pareto-optimal solutions for dynamic time allocation in multi-function radar systems under time budget constraints. 
Our approach succeeds in finding diverse, high-quality solutions across the objective space,
outperforming a heuristic equal allocation approach. Our results demonstrate that balanced exploration is crucial for discovering diverse solutions in multi-objective optimization. 
The CDRL framework offers radar operators a flexible tool to dynamically adjust trade-offs between different tasks by adjusting the parameters in the proposed learning framework, enabling real-time optimization of resource allocation in response to dynamic task priorities.

\newpage
\bibliographystyle{IEEEtran}
\bibliography{references}

\end{document}